%% file: main.tex
\documentclass[lettersize,journal]{IEEEtran}
\usepackage{amsmath,amsfonts}
\usepackage{algorithm}
\usepackage{array}
\usepackage{textcomp}
\usepackage{stfloats}
\usepackage{url}
\usepackage{verbatim}
\usepackage{graphicx}
\usepackage{cite}
\usepackage{soul}
\usepackage{algorithmicx}
\usepackage{algpseudocode}

\usepackage{bibentry}
\usepackage{subfigure}
\usepackage{color}
\usepackage{diagbox}
\usepackage{listings}

\usepackage[figuresright]{rotating}

\begin{document}

\title{AdaNAS: Adaptively Post-processing with Self-supervised Neural Architecture Search for Ensemble Rainfall Forecasts}

\author{
Yingpeng Wen, Weijiang Yu*, Fudan Zheng, Dan Huang, Nong Xiao*\\
~\IEEEmembership{Sun Yat-Sen University, School of Computer Science}
\thanks{The authors are with the Sun Yat-Sen University, School of Computer Science ,
Guangzhou, Guangdong, 511400, CN (e-mail:weijiangyu8@gmail.com). * are corresponding authors}}

\markboth{Journal of \LaTeX\ Class Files,~Vol.~14, No.~8, August~2021}%
{Shell \MakeLowercase{\textit{et al.}}: A Sample Article Using IEEEtran.cls for IEEE Journals}


\maketitle

\begin{abstract}

Previous post-processing studies on rainfall forecasts using numerical weather prediction (NWP) mainly focus on statistics-based aspects, while learning-based aspects are rarely investigated. Although some manually-designed models are proposed to raise accuracy, they are customized networks, which need to be repeatedly tried and verified, at a huge cost in time and labor. Therefore, a self-supervised neural architecture search (NAS) method without significant manual efforts called AdaNAS is proposed in this study to perform rainfall forecast post-processing and predict rainfall with high accuracy. In addition, we design a rainfall-aware search space to significantly improve forecasts for high-rainfall areas. Furthermore, we propose a rainfall-level regularization function to eliminate the effect of noise data during the training. Validation experiments have been performed under the cases of \emph{None}, \emph{Light}, \emph{Moderate}, \emph{Heavy} and \emph{Violent} on a large-scale precipitation benchmark named TIGGE. Finally, the average mean-absolute error (MAE) and average root-mean-square error (RMSE) of the proposed AdaNAS model are 0.98 and 2.04 mm/day, respectively. Additionally, the proposed AdaNAS model is compared with other neural architecture search methods and previous studies. Compared results reveal the satisfactory performance and superiority of the proposed AdaNAS model in terms of precipitation amount prediction and intensity classification. Concretely, the proposed AdaNAS model outperformed previous best-performing manual methods with MAE and RMSE improving by 80.5\% and 80.3\%, respectively.
\end{abstract}

\begin{IEEEkeywords}
Rainfall Forecasts, Precipitation Prediction, Automated Machine Learning, Neural Architecture Search, Self-supervised Learning,  and Rainfall-level Regularization.
\end{IEEEkeywords}

\input{Introduction}
\input{StudyareaandCriteria.tex}

\input{Method}

\input{Experiments}

\input{Conclusion}
\input{Acknowledgements}
\input{Appendices}


\bibliographystyle{IEEEtran}
\bibliography{ref}

\newpage

\vfill

\end{document}

%% file: Introduction.tex
\section{Introduction}
\IEEEPARstart{H}{eavy} precipitation\footnote{The rainfall is also regarded as precipitation in our paper.} events can cause severe flooding and further result in economic damage and loss of life \cite{zhao2021hourly}. Accurate and quantitative forecasting of rainfall help develop effective measures and prevent loss of life and property from floods and landslides. The previous technologies for rainfall forecasting mainly depend on statistic-based paradigms, like the numerical weather prediction (NWP) model systems~\cite{michalakes2001development,tang2013benefits,seity2011arome},  ensemble prediction systems~\cite{ebert2001ability,zhi2011multi} and post-processing techniques~\cite{dai2016situation}. Since individual NWP model runs and ensemble systems are subject to biases and dispersion errors, their predictions can be improved by statistical post-processing techniques\cite{cuo2011review,gneiting2014probabilistic,schaake2007hepex}. The traditional post-processing processes use simple operations to synthesize ensemble forecasts. F. Kong et al. propose the ensemble mean (EM) method~\cite{kong2006multiresolution} to average multiple ensemble forecasts to predict precipitation. The probability matching (PM) method~\cite{ebert2001ability} redistributes the precipitation rates in the ensemble mean by the distribution of precipitation rates from the available QPFs. The best percentile (BP) method~\cite{dai2016situation} adopts asymmetric normal distribution for parameter estimation to calculate probabilistic quantitative precipitation and percentile forecasts. X. Zhi et al. establish the weighted bias-removed ensemble mean (WEM)~\cite{zhi2011multi} to predict precipitation based on the weighted average of ensemble forecasts, where the weights are calculated from the historical errors of the ensemble forecasts. Although these methods exhibit different advantages in a variety of test metrics, accuracy is still hardly commendable.

\begin{figure}
	\centering
	\includegraphics[width=1.0\linewidth]{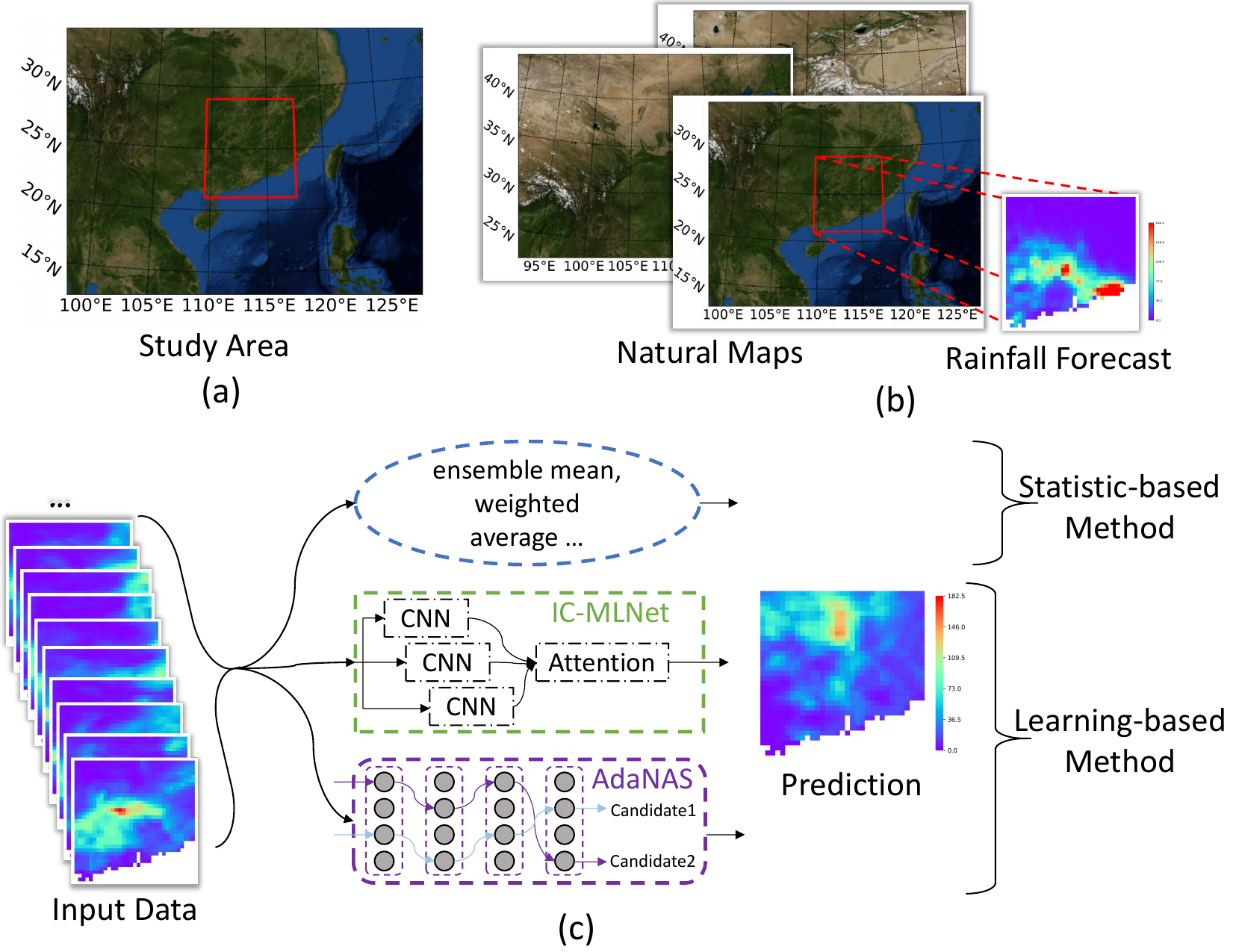}
	\caption{(a) The study area $[21.0^{o}N \sim 29.0^{o}N, 109.5^{o}E \sim 117.5^{o}E]$ includes part of the coastal and inland regions of southern China. (b) The data characteristic of rainfall forecast are abstracted from the raw observational data by NWP models. (c) Overview of previous methods applied to rainfall forecast and our AdaNAS, which can be summarized by two aspects: statistic-based method and learning-based method.
	}
	\label{fig:intro}
\end{figure}

Machine learning methods have been applied to the post-processing of precipitation forecasts in recent years. The advantages of machine learning methods include incorporating diverse features and automatically extracting useful information \cite{zhang2021machine}. Machine learning-based methods also help to model complex relationships between predictors and predictands to improve forecasting skills \cite{li2022convolutional}. W. Li et al. \cite{li2022convolutional} develop a convolutional neural network (CNN)-based post-processing method for precipitation forecasts with spatial information and atmospheric circulation variables. M. Ghazvinian et al. propose a hybrid artificial neural network (ANN) that uses the censored, shifted gamma distribution (CSGD) as the predictive distribution and uses (ANN) to estimate the distributional parameters of CSGD \cite{ghazvinian2021novel}. F. Xu et al. design a multi-layer network (IC-MLNet)-based post-processing method~\cite{xu2021multi} to predict precipitation, showing better performance than traditional methods \cite{kong2006multiresolution,ebert2001ability,dai2016situation,zhi2011multi} in precipitation amount and precipitation levels prediction. For more details on the available post-processing models, the reader is referred to the relevant books and reviews \cite{duan2019handbook,vannitsem2021statistical,vannitsem2018statistical}.

Although these learning-based studies perform out of statistic-based methods, it is time-consuming to trial and error manually designed networks. Besides, the current learning-based methods are imprisoned due to the high requirement of expert domain knowledge to design the network. These shortcomings of manually designed neural networks have hindered the development and application of post-processing technology for precipitation forecasts. 
Therefore, we aim to develop efficient learning-based methods for automatically designing networks for the complex post-processing of ensemble rainfall forecasts.

Neural architecture search (NAS) aims to search for a robust and well-performing neural architecture by selecting and combining various basic operations from a predefined search space \cite{he2021automl}. For computer vision tasks, various NAS methods~\cite{xu2019pc,he2020milenas} have emerged including evolutionary-based~\cite{ahmed2018maskconnect,you2020greedynas}, reinforcement learning (RL)-based~\cite{cai2018proxylessnas,tan2019efficientnet} and gradient descent (GD)-based~\cite{liu2018darts,he2020milenas,xu2019pc} NAS methods. The differentiable NAS optimized by the gradient descent method is favored for its fast search speed. Most of the gradient descent-based neural network architecture search methods are dedicated to computer vision research, such as MileNAS~\cite{he2020milenas}, DARTS\cite{liu2018darts} and PC-DARTS~\cite{xu2019pc}.

In this paper, a model based on the adaptively self-supervised neural architecture search algorithm called AdaNAS is first proposed to ensemble rainfall forecast post-processing with high accuracy, which can effectively generate deterministic precipitation as well as reduce lots of manual efforts. The self-supervised technology is applied in the neural network architecture search method to make fuller use of the limited data samples. In addition, we customize a rainfall-aware search space to significantly improve forecasting in high-rainfall areas. Since the search space consisting of the residual block (RB) is not successful in predicting heavy rainfall, we design space-aware block (SAB) and channel-aware (CAB) to improve the accuracy of heavy rainfall prediction. Furthermore, a rainfall-level regularization function is presented to eliminate the effect of noise data in the training. The performance of the AdaNAS model is evaluated and the results reveal that the proposed model is superior to previous best-performing manual methods and other NAS methods in terms of precipitation amount prediction and intensity classification.

%% file: StudyareaandCriteria.tex
\section{STUDY AREA AND CRITERIA}
This section describes the details of the data and study area in Section~\ref{sec:data}. In Section~\ref{sec:crateria}, we describe the evaluation criteria for precipitation prediction. 
\subsection{Study Area and Data Description}
\label{sec:data}
\subsubsection{Study Area}
The study area $[21.0^{o}N \sim 29.0^{o}N, 109.5^{o}E \sim 117.5^{o}E]$ is in southern China, including part of the coastal and inland region, as shown in Fig.~\ref{fig:intro} (a). 
The northern and central region is characterized by a subtropical monsoon climate, while the southern region is a tropical monsoon climate. The annual rainfall in the study area is approximately 1500 mm annually. The rainy season typically occurs from June to October. Extreme weather events such as typhoons are frequent in summer in the coastal region, increasing the difficulty of rainfall forecasting. Therefore, one of the main challenges in the study area is the rainfall forecasting of heavy coastal rainfall.

\subsubsection{Ensemble Forecast Data}
We use ensemble data from the TIGGE\footnote{https://apps.ecmwf.int/datasets/data/tigge/} dataset, a classic rainfall prediction dataset used in several works~\cite{xu2021multi,zhi2011multi}. To demonstrate that our post-processing method can be well adapted to ensemble forecasting models (single-model and multi-model ensemble forecasting), our input data are divided into single-model datasets (Smod) and multi-model datasets (Mmod). The single-model dataset is the ensemble forecast from ECMWF Center, generated by 50 random initial conditions, and the multi-model dataset is the deterministic forecast from NWP model system in 4 centers, namely UKMO, NCEP, JMA and ECMWF. 
TIGGE issues daily weather forecasts for 366 hours at UTC0000 and UTC1200, but only $6 \sim 30, 12 \sim 36, 18 \sim 42$ and $24 \sim 48$ forecast hours are used in these datasets.
\subsubsection{Observation Data}

We collect observation data from 7247 automatic stations\footnote{http://data.cma.cn/en/?r=data/detail\&dataCode=A.0012.0001} in the region $[21.0^{o}N \sim 29.0^{o}N, 109.5^{o}E \sim 117.5^{o}E]$. As there are many missing data in the observed data in 2014, we only use forecasts and observations in 2013, 2015 and 2016. We use the mean interpolation method to make the observed data format consistent with the ensemble forecast, where the ensemble forecast data are the input data as shown in Fig.~\ref{fig:intro} (b) and the observation data are the labels. We strictly screen the data to ensure their availability. The following three types of data are excluded: observations and ensemble forecasts that do not correspond to each other in time, those missing one or more ensemble members, and those with all zero values for either ensemble members or observations. The remaining single-model data contains 4160 samples and the multi-model data contains 3785 samples. 

\subsubsection{Training Dataset and Validation Dataset}
We divide the training dataset and validation dataset by timeline at a ratio of 9:1, which is the same as in IC-MLNet \cite{xu2021multi} does. For the validation dataset to adequately verify the effectiveness of the post-processing method, the distribution of precipitation values should be consistent with the entire dataset. Thus, we chose the validation data from 2016-04-16 to 2016-06-13, which have a precipitation distribution closer to the overall data (2 winters and 3 flood seasons).

\subsection{Evalutaion Criteria}
\label{sec:crateria}

\begin{table}
\caption{multi-category contingency table}
\label{contingency}
\centering
\begin{tabular}{|c|cccc|c|}
\hline
\diagbox{i}{j} & 1 & 2 & $\cdot\cdot\cdot$ & L & Total \\
\hline
1 & $n_{1,1}$ & $n_{1,2}$ & $\cdot\cdot\cdot$ & $n_{1,L}$ & $N'_1$ \\

2 & $n_{2,1}$ & $n_{1,2}$ & $\cdot\cdot\cdot$ & $n_{2,L}$ & $N'_2$ \\

$\cdot\cdot\cdot$ & $\cdot\cdot\cdot$ & $\cdot\cdot\cdot$ & $\cdot\cdot\cdot$ & $\cdot\cdot\cdot$ & $\cdot\cdot\cdot$ \\

L & $n_{L,1}$ & $n_{L,2}$ & $\cdot\cdot\cdot$ & $n_{L,L}$ & $N'_L$ \\
\hline
Total & $N_1$ & $N_2$ & $\cdot\cdot\cdot$ & $N_L$ & $N_{T}$ \\
\hline
\end{tabular}

\end{table}

\begin{equation}
\label{bias}
Bias = \sum_{i=1}^n \tilde{y}_i \bigg/ \sum_{i=1}^n y_i,
\end{equation}
To fully demonstrate the outstanding performance of our post-processing methods, we use several evaluation criteria as in IC-MLNet \cite{xu2021multi}, including \emph{bias}, \emph{mean absolute error} (MAE), \emph{root mean square error} (RMSE), \emph{Nash-Sutcliffe model efficiency} coefficient (NSE), precipitation classification \emph{Accuracy} (ACC) and \emph{Heidke skill score} (HSS). \emph{Bias} is the ratio of the prediction sum as Equation (\ref{bias}) to the label sum and is used to measure the overall deviation of the prediction from the label. The perfect score of \emph{bias} is 1 when there is no deviation of the prediction sum from the label sum. MAE and RMSE are routinely used to measure the distance between predictions and labels, ranging from 0 to positive infinity, with a perfect score of 0, when predictions are equal to labels. The specific equations for MAE and RMSE are so commonly used that they are no longer provided here. \emph{Nash-Sutcliffe model efficiency coefficient} (NSE) reflects the numerical relationship between the variance of the prediction error and the variance of the observed data as Equation (\ref{NSE}), which is used to assess the predictive skill of hydrological models. In the case of a perfect model with zero estimation error variance, the obtained $NSE = 1$. 
\begin{equation}
\label{NSE}
NSE = 1 - \sum_{i=1}^n ||\tilde{y}_i - y_i ||^2_2 \bigg/ \sum_{i=1}^n || y_i - \bar{y}_i||_2^2,
\end{equation}
where $\tilde{y}$ denotes the predicted value, $y$ denotes the observed value (labels), and $\bar{y}$ denotes the mean of the observed value.

We introduce rainfall prediction ACC and HSS to demonstrate whether it can correctly predict the level of rainfall to verify the practical applicability of our post-processing method since the weather center does not need to report specific precipitation values but reports the level of rainfall when reporting forecast results to the public. We classify rainfall into five categories according to its magnitude: 1. \emph{None} $[0.0, 0.1)mm/day$, 2. \emph{Light} $[0.1, 10.1)mm/day$, 3. \emph{Moderate} $[10.1, 25.1)mm/day$, 4. \emph{Heavy} $[25.1, 50.1)mm/day$ and 5. \emph{Violent} $[50.1, \infty)mm/day$. We create a multi-category contingency table according to the levels for all forecast maps as Table \ref{contingency}. $L=5$ is the number of rainfall categories; $n_{i,j}$ denotes the number of cases where the observation level i is predicted to be level j; $N'_i$ denotes the number of rainfall predicted to be level i; $N_j$ represents the number of observations to be level j, and $N_T$ denotes the total number of predictions.

Based on this table we can calculate the following evaluation criteria: ACC and HSS. ACC measures how well the predicted and observed data rainfall levels match, as shown in Equation (\ref{Acc}), ranging from 0 to 1. In the perfect case, all predicted levels are the same as observed levels, and ACC is 1. ACC reflects the percentage of correct predictions, but the uneven distribution of the 5-category sample affects ACC's ability to reflect the real situation. Therefore, we also introduce another evaluation criterion, HSS as Equation (\ref{HSS}), which can exclude correct predictions due to random chance and obtain a more realistic percentage of correct predictions. HSS ranges from negative infinity to 1 and the perfect score is 1.

\begin{equation}
\label{Acc}
ACC = \frac{1}{N_T}\sum_{i=1}^L n_{i,i},
\end{equation}

\begin{equation}
\label{HSS}
HSS = \frac{\frac{1}{N_T}\sum\limits_{i=1}^L n_{i,i} - \frac{1}{N_T^2}\sum\limits_{i=1}^L N'_i N_i}{1 - \frac{1}{ N_T^2}\sum\limits_{i=1}^L N'_i N_i}.
\end{equation}

%% file: Method.tex
\section{Method Description}
In this section, we describe the task and the post-processing methods used in the research. Firstly, we describe the precipitation post-processing problem In Section \ref{sec:problem}. Section \ref{sec:NAS} introduces our proposed adaptive neural architecture search method, including self-supervised search, search space and regularization function.

\subsection{Problem Definition}
\label{sec:problem}
\begin{figure}[t]
\begin{center}
\includegraphics[width=0.8\linewidth]{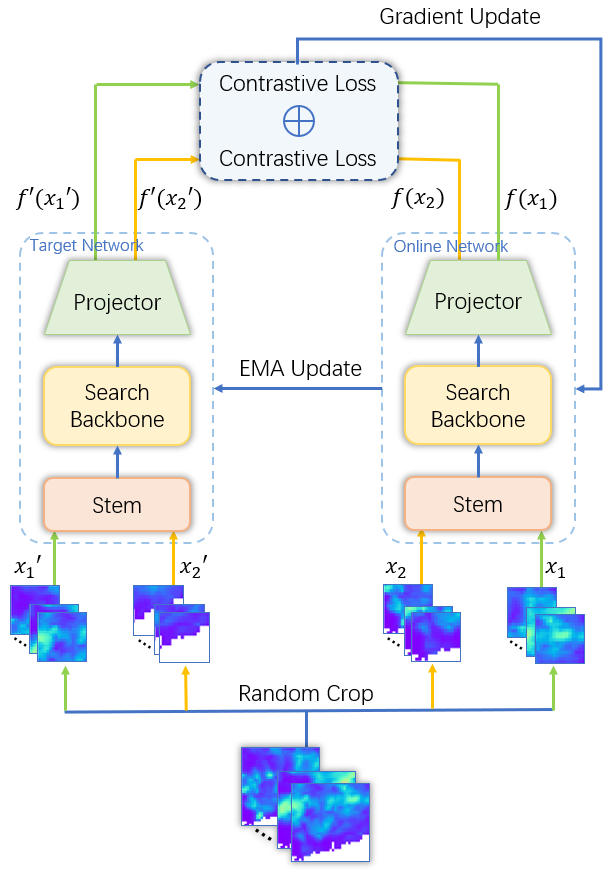}
\end{center}
   \caption{The overview of our AdaNas by using contrastive learning to adaptively search suitable architecture for ensemble rainfall forecast post-processing.}
\label{fig:framework}
\end{figure}

The ensemble forecast post-processing is aim to generate an accurate deterministic forecast based on the ensemble forecast data, i.e. average rainfall for the next 24h predicted by the NWP model. In other words, we integrate the ensemble forecast data generated from multiple NWP models or a single NWP model with multiple initial conditions, to output the average rainfall forecast for the next 24h in the same area.
The post-processing process can be expressed as follows:
\begin{equation}
y = f(x;W),
\end{equation}
where $x \in R^{c\times w \times h}$ denotes the ensemble forecast data; $f$ denotes the forecast model; $W$ is the model parameter; $y \in R^{w \times h}$ is the deterministic forecast; $w = 33$ and $h = 33$ are respectively the width and height of the region after gridding; $c$ denotes the number of channels, determined by the number of NWP models and the number of random initial states; $c$ is 50 and 4 in Smod and Mmod respectively.

\vspace{-1mm}
\subsection{Adaptive Neural Architecture Search}
\label{sec:NAS}

We propose an advanced AdaNAS method to design suitable network architectures automatically, avoiding a mass of manual effort and achieving excellent performance. In summary, our approach is divided into two steps, including searching for architecture and training the searched model. To better adapt the AdaNAS to the rainfall forecast, we present a self-supervised search strategy, a rainfall-aware search space and a rainfall-level regularization function.

\vspace{-1mm}
\subsubsection{Self-supervised Search}
We adopt a block-wise self-supervised comparative learning approach to search the network architecture, instead of using shared weights as in one-shot NAS. To reduce the computational effort, the weight-sharing rating scheme in the one-shot NAS method is adopted by most NAS methods. However, the architecture ranking estimated with shared weights is not necessarily the true architecture ranking because there must be a huge gap between the shared weights and the optimal weights of sub-networks. Some studies~\cite{yu2019evaluating,yang2019evaluation,zela2020bench} pointed out that the evaluation method of shared weights has low accuracy. In addition, some theoretical and empirical studies~\cite{li2020improving,moons2020distilling,li2020block} demonstrated that reducing shared weights can effectively improve the accuracy of evaluating architecture ranking.

A block-wise method is ideal to reduce shared weights by splitting the depth of the network, keeping the original search space and solving the dilemma of shared weights. Each block of the hyper-network is trained separately before being connected to the overall search. Thus, each of our blocks has a separate structure, instead of being stacked with the same structure, which makes our network architecture more malleable.


The outline of the neural network architecture search algorithm is shown in Algorithm \ref{alg:Alogrithm}. It uses a self-supervised contrastive learning approach to update the architecture weights $\theta$ and model weights $W$ separately, and ultimately outputs the optimal architecture $a^*$, where $u$ is a factor to balance the update frequencies of $\theta$ and $W$; $T$ denotes the number of epochs for search process; $N$ denotes the number of building block; $A$ is a collection of architectures.


\begin{algorithm}[H]
\footnotesize
\caption{Search Algorithm for AdaNAS} 
{\bf Input:}  Dataset $D$  \\
{\bf Initialization:} Random initialize $\theta$ and $W$; 
\begin{algorithmic}

\For{$t = 1$ to $T$} 
\If{$t$ \% $u$ \textgreater 0}
\State Random sample a mini-batch $d$ $\subseteq$ $D$; 
\State Sample $a_1,a_2$ $\sim$A($\theta$) with respect to $\theta$;
\State Update $W$ by descending $\nabla_{W}L(a_1,a_2,W)$ on $d$;
\Else
\State i = t // (T // N) + 1;
\State Random sample a mini-batch d $\subseteq$ $D$;
\State Sample architectures $a_1,a_2$ $\sim$A($\theta$) with respect 
\State to $\theta$ by picking operations in i-th building block;
\State Update $\theta$ by descending $\nabla_{\theta}L(a_1,a_2,W)$ on $d$;
\EndIf
\EndFor
\State
\Return a* by picking operations with the largest value in $\theta$* for each block.
\end{algorithmic}
\label{alg:Alogrithm}
\end{algorithm}

Self-supervised comparative learning uses auxiliary tasks to mine information from unsupervised data for improving the quality of downstream tasks, i.e., rainfall forecast.
Concretely, we replicate a target network that has the same architecture as the online network. As shown in Figure~\ref{fig:framework}, the initialized parameters of the target network are the same as those of the online network. The exponential moving average (EMA)~\cite{He2020MomentumCF} update is done during the search according to the parameters of the online network. Whenever data is input, it will be randomly cropped into four seeds $x_1,x_2,x'_1$ and $x'_2$, two for each network input. The backbone of the whole architecture contains multiple blocks, and $x_1$ and $x_2$ will choose different paths when passing through the backbone to search for the optimal architecture. The online network performs a gradient update~\cite{rumelhart1986learning} according to the contrastive loss of output data. The pseudo-code of the search process is shown as Algorithm \ref{pseudocode}.

\begin{algorithm}[t]
\caption{Pseudocode of AdaNAS searching in a PyTorch-like style.}
\label{pseudocode}
\definecolor{codeblue}{rgb}{0.25,0.5,0.75}
\definecolor{codekw}{rgb}{0.85, 0.18, 0.50}
\vspace{-1mm}
\begin{lstlisting}[language=python]
# f_o, f_t: online network and target network
# m: momentum
f_t.params = f_o.params # initialize
# load a minibatch x with N samples
for x in loader:
  x_1, x_2, x_1', x_2' = crop(x) # randomly crop x
  # online network for queries
  q_1, q_2 = f_o(x_1), f_o(x_2)
  # target network for keys
  k_1, k_2 = f_t(x_1'), f_t(x_2')
  
  # contrastive loss
  loss = MSEloss(q_1, k_1) + MSEloss(q_2, k_2)
  # backward propagation update
  loss.backward()
  update(f_q.params)
  
  # momentum update (EMA)
  f_t.params = m*f_t.params + (1-m)*f_o.params
\end{lstlisting}
\end{algorithm}
\subsubsection{Search Space}
Search space is an important part of the NAS, which determines the search scope of the NAS. Inspired by the existing works~\cite{Resnet,yang2021simam,vaswani17nips}, we design our search space consisting of suitable CNNs and transformer operations, including residual block (RB), space-aware block (SAB) and channel-aware block (CAB). RB is the main structure of Resnet~\cite{Resnet} and is used to extract features. SAB and CAB are transformer operations that focus attention on pixels with salient features. 

\begin{figure}
\centering
\includegraphics[width=0.45\textwidth]{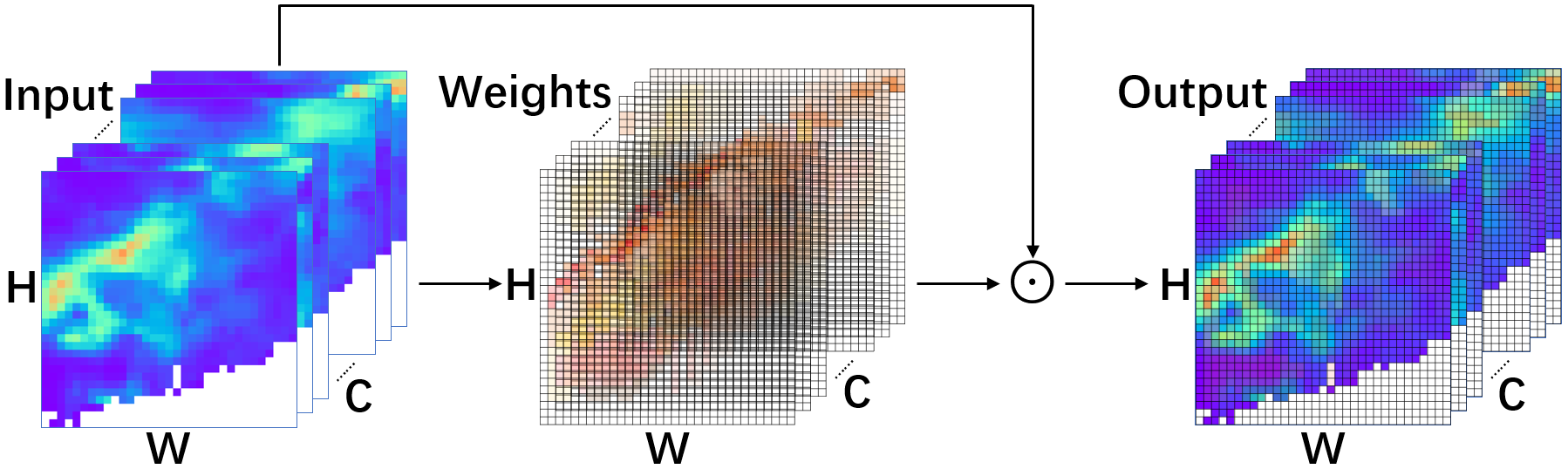}
\caption{Channel-aware block. Channel-aware block focuses on high rainfall pixels and assigns large weights to highlight the features of these high rainfall pixels}
\label{fig:CAB_3}
\end{figure}

\begin{figure}
	\centering
	\includegraphics[width=1.0\linewidth]{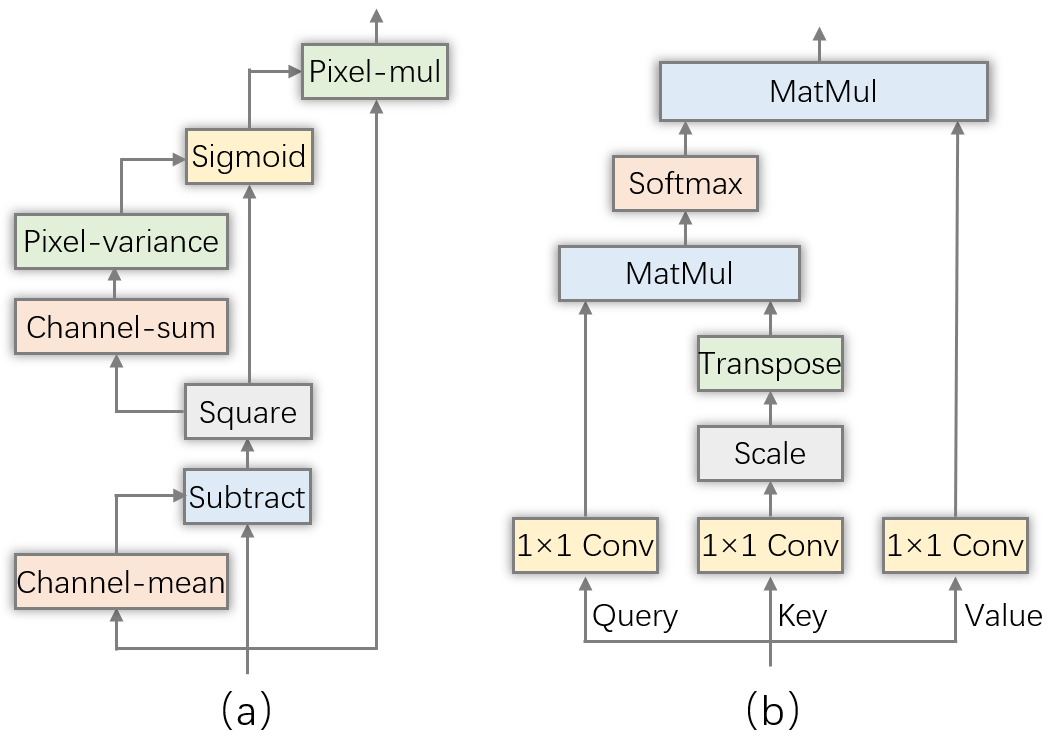}
	\caption{The details of our channel-aware block (a) and space-aware block (b).}
	\label{fig:operations}
\end{figure}

\begin{figure}
\centering
\includegraphics[width=0.5\textwidth]{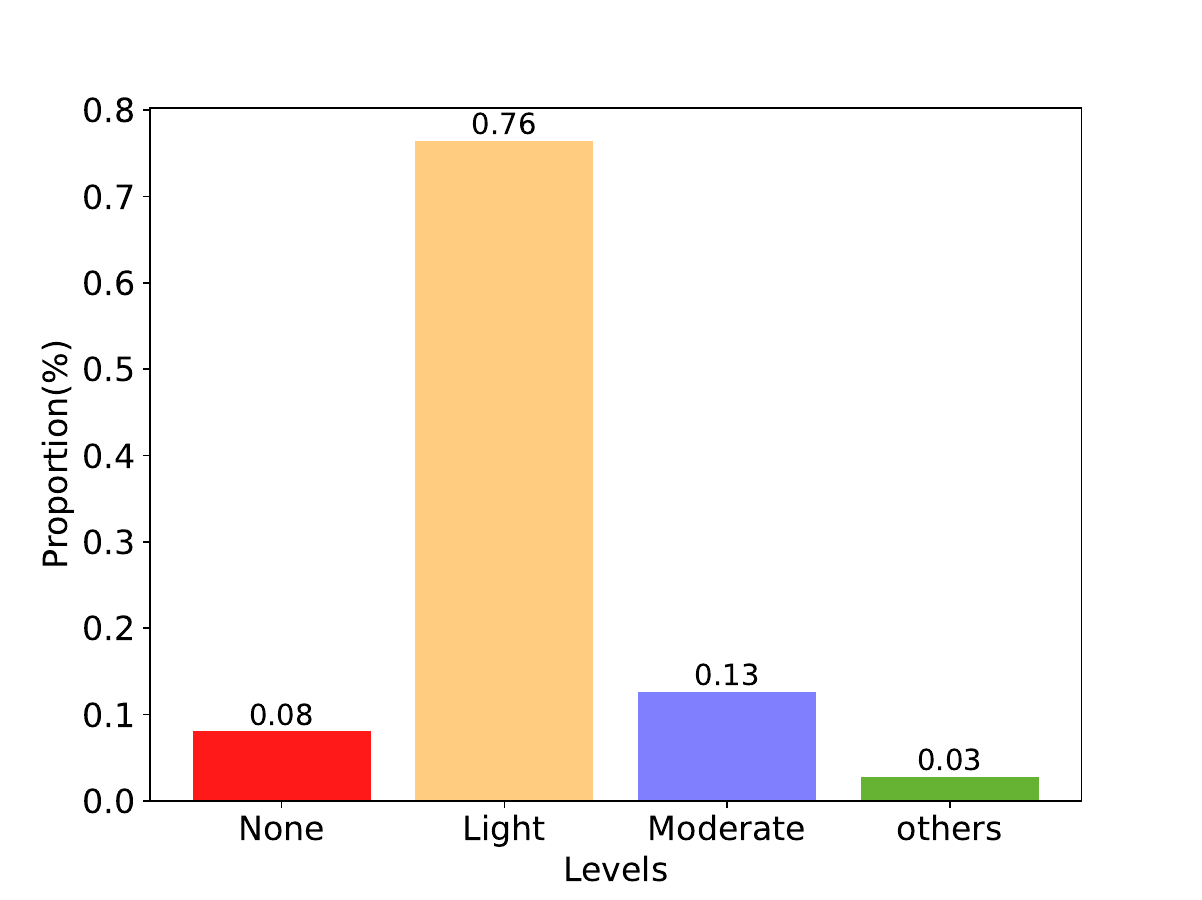}
\caption{Proportion of precipitation events in multi-model dataset. The proportion of rainfall at different levels is extremely uneven, with Light accounting for nearly 80\%.}
\label{distribution}
\end{figure}

To ensure a more accurate prediction of the less frequent cases such as heavy rainfall, we introduce the transformer-based blocks CAB and SAB. As illustrated in Figure~\ref{distribution}, the proportion of Light and None accounts for more than 80\%, resulting in difficulty to predict heavy rain. CAB and SAB are used to capture the heavy rainfall pixels in the ensemble forecast data and assign a greater weight to them to highlight outstanding features. This enables the model to focus on these heavy rainfall pixels as shown in Figure \ref{fig:CAB_3} and to predict the heavy rainfall areas more accurately. In Figure \ref{fig:mae}, the visualization of rainfall distribution also demonstrates that our method obtains significant results in coastal areas with heavy rainfall. As a more robust transformer-based block, the channel-aware block determines the importance of a neuron by its inhibitory effect on the surrounding space, with the following expression:
\begin{equation}
z' = z \bigodot sigmoid(\frac{(z - \bar{z})^2}{4 \times (\frac{sum((z - \bar{z})^2)}{n} + \lambda)} + 0.5),
\end{equation}
where $z$ denotes the input data; $\bar{z}$ denotes the expectation of $z$; $sum(\cdot)$ denotes the internal summation; $n = w \times h - 1$, $w$ and $h$ denote the width and height of input data; $\lambda=10^{-4}$ is hyperparameter. $sigmoid(\frac{(z - \bar{z})^2}{4 \times (\frac{sum((z - \bar{z})^2)}{n} + \lambda)} + 0.5)$ is equivalent to the weight given to a pixel by CAB based on the difference between the pixel points, where pixels with significant differences are given a higher weight. The more detailed pipeline of our SAB and CAB can be seen in Figure~\ref{fig:operations}.

\subsubsection{Retrain Process}
We use the self-supervised search method to search for a suitable neural network model in the search space we have designed. The searched neural architecture is shown in Figure~\ref{fig:search_backbone}. This model mainly includes a stem layer, a search backbone and a projector. The stem layer includes a convolutional layer, a batch normalization layer, an active layer and a max-pooling layer. The projector includes a pooling layer and a fully connected layer. The model is trained fully supervised on the dataset with the input being the ensemble forecast data and the labels being the observations. Different from the normal fully supervised training, we design a novel regularization equation to improve the accuracy.

\subsubsection{Regularization Function}
The regression tasks like rainfall prediction usually use a distance metric such as \emph{mean square error} (MSE) as a regularization function. However, the MSE is a constraint term for normal rainfall prediction rather than the fine-grained rainfall level classification. Although existing methods optimized by the MSE can predict some examples, it still restrains the performance of the prediction in rainfall levels due to the lack of effective constraints. Therefore, we design a regularization function by introducing a categorical criterion, \emph{Heidke skill score} (HSS), to improve the accuracy of the rainfall level classification. HSS is a categorical criterion that measures the accuracy of prediction by excluding the case where the random prediction is correct, as shown in Equation \ref{HSS}. The reason we introduce HSS instead of \emph{Accuracy} (ACC) is that the general ACC scores cannot truly reflect the predictive power of the model on a dataset with a severely uneven distribution of categories. For example, the percentage of \emph{light} precipitation events reaches 76.4\% (As shown in Figure~\ref{distribution}, \emph{light}, \emph{moderate}, \emph{none} and others precipitation levels account for 76.4\%, 12.7\%, 8.1\% and 2.4\% in Mmod, respectively.). The model only needs to predict \emph{light} for any input to guarantee an ACC of 76.4\%, which is significantly higher than all the models in Table \ref{main table}. However, such a model does not have qualified predictive power. HSS rules out this possibility, so we introduce the HSS into the regularization function with the following expression:
\begin{equation}
Loss = loss_{MSE} + \frac{c_H}{\max(loss_{HSS},\epsilon)},
\end{equation}
where $loss_{MSE}$ and $loss_{HSS}$ are the loss of precipitation and intensity classification respectively; $c_H$ is a coefficient; $\epsilon$ is a tiny constant, here we take $10^{-10}$. Experiments \ref{sec:loss} show that the performance of the regularization function introduced with $loss_{HSS}$ is improved notably, and its performance is affected by coefficients $c_H$.


%% file: Experiments.tex
\section{Experiments}
\label{sec: PAP}

\begin{table*}
\caption{Results of precipitation amounts and intensity classification on Smod and Mmod.}
\label{main table}
\footnotesize
\centering
\setlength\tabcolsep{3pt}
\begin{tabular}{|c|llll|ll|llll|ll|}
\hline
Dataset & \multicolumn{6}{|c|}{Smod} & \multicolumn{6}{|c|}{Mmod} \\
\hline
 & \multicolumn{4}{|c|}{Precipitation amounts}& \multicolumn{2}{|c|}{levels} & \multicolumn{4}{|c|}{Precipitation amounts}& \multicolumn{2}{|c|}{levels} \\
\hline
\diagbox{Method}{Criterion} & Bias & MAE &RMSE &NSE &ACC &HSS & Bias & MAE &RMSE &NSE &ACC &HSS \\
\hline
EM & 1.2733 & 6.1073 &11.5020 &0.1971 &0.5330 &0.2653 & 1.2991 & 7.1607& 12.5314 & 0.1891& 0.5041& 0.2528 \\
PM & 1.2794 & 6.4901 &13.2375 &-0.0635 & 0.5616& 0.3179 & 1.3006 & 7.5511& 14.2221 &-0.0445& 0.5247& 0.2865 \\
BP & 1.5740 & 10.1254 &24.2992 &-2.5834 &0.5852 &0.3525 & 1.2714 & 8.8641& 18.6596& -0.7980& 0.5593& 0.3238 \\
WEM & 1.2497 &  6.8066 &44.9405 &-11.2570 &0.5559& 0.3407 & 1.3767 & 8.9343& 39.4014& -7.0168& 0.5027& 0.2817 \\
\hline
IC-MLNet & \emph{0.8176} & \emph{5.0185} &\emph{10.3863} &\emph{0.3453} &\emph{0.6030}& \emph{0.3914} &  \emph{0.7899} &  \emph{5.8773}& \emph{11.3828}& \emph{0.3246}& \emph{0.5784}& \emph{0.3681} \\
\hline
MiLeNas & 1.1832&	1.2707&	2.2244&	0.2231&	0.6146&	0.2366 & 1.0205&	2.8079&	5.6283&	0.3318&	0.6190	&0.3303 \\
PC-DARTS & 0.8250&	1.0213&	2.0615&	0.0759&	0.6512&	0.3082 & 1.0296	&2.8553&	5.7589&	0.2306&	0.6295&	0.3344 \\
NSAS & 1.1605&	1.1534&	2.1528&	0.1652&	0.6774&	0.3695 & 0.9664	&2.6950&	5.5164&	0.3009&	0.6072&	0.2748 \\
\hline
AdaNAS-random & 0.8482&	1.0961&	2.2196&	0.4098	&0.6776	&0.3783 & 1.0777&	2.7685&	6.0750&	0.4495&	0.6257&	0.3416 \\
AdaNAS-mse & \textbf{0.8695}	&0.9917&	2.0754&	0.4305&	0.6916&	0.3930 & \textbf{0.9874}&	2.6507&	5.3714&	0.5062&	0.6403&	0.3635 \\
AdaNAS-hss & 0.8618&	\textbf{0.9796}&	\textbf{2.0420}&	\textbf{0.4460}&	\textbf{0.7064}&	\textbf{0.4236} & 0.9833 &	\textbf{2.6280}&	\textbf{5.3217}&	\textbf{0.5180}&	\textbf{0.6457}&	\textbf{0.3723} \\
\hline
\end{tabular}

\end{table*}


We conduct experiments and analysis on six aspects, namely precipitation amount prediction, precipitation intensity classification, regularized equation analysis, predictive distribution analysis, ablation study and time consumption analysis. Some important hyperparameter configurations are shown in Appendix \ref{secA1}.
\subsection{Precipitation Amounts Prediction}

\begin{figure*}[t]
\begin{center}
\includegraphics[width=1.0\linewidth]{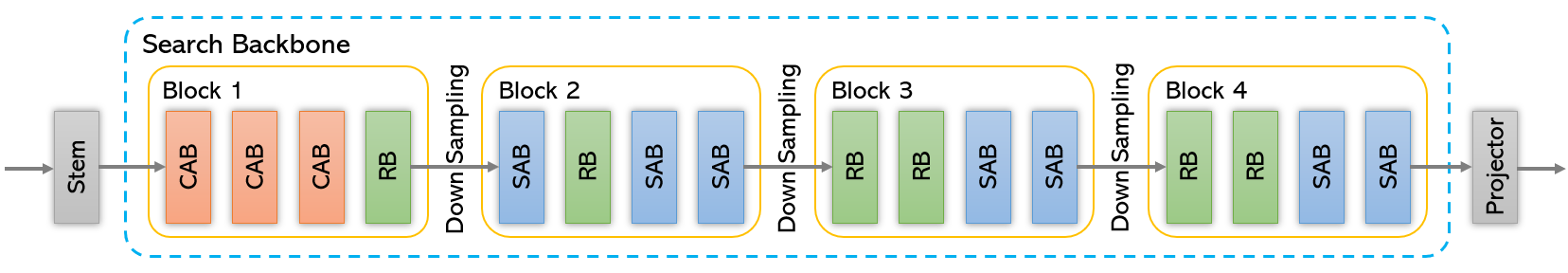}
\end{center}
    \caption{The detailed search architecture of our AdaNAS.}
\label{fig:search_backbone}
\end{figure*}

AdaNAS-hss achieves the best performance in all rainfall evaluation criteria as shown in Table \ref{main table}, where the regularization function of AdaNAS-hss includes HSS while AdaNAS-mse does not. Our model improves 5.4\%, 80.5\%, 80.3\% and 29.2\% in Bias, MAE, RMSE and NSE, respectively, on Smod compared with the previous best-performing IC-MLNet. Consistently, our model improves 24.5\%, 55.3\%, 53.2\% and 59.6\% in Bias, MAE, RMSE and NSE, respectively, on Mmod compared with the previous best-performing IC-MLNet. The auto-designed architecture, as shown in Figure \ref{fig:search_backbone}, consists of four blocks and three operations.

Our AdaNAS also outperforms other NAS methods (MiLeNas, PC-DARTS and NSAS~\cite{NSAS}) customized by normal RGB images. It is significantly improved in NSE compared to MiLeNAS, PC-DARTS and NSAS with more than 56.1\% improvement on both Smod and Mmod. 
In Table \ref{main table}, all NAS methods have excellent performance in Bias, MAE and RMSE, outperforming IC-MLNet. However, it shows a clear deficiency in the NSE metrics. These results indicate that the NAS approach has a natural and significant advantage in Bias, MAE and RMSE metrics, but this advantage is lost in the NSE. In addition, the comparison of AdaNAS-mse and AdaNAS-hss shows that introducing HSS into the regularization equation improves MAE, RMSE and NSE scores on both datasets, while Bias decreases a bit.


To prove statistically that our method is indeed more advantageous than the other methods, we use the Diebold-Mariano test \cite{diebold1995paring} to perform multiple tests on our method and the other methods. The experimental results shown in TABLE \ref{table: DM} prove that there is indeed a statistical difference between the results of our predictions and those of the other methods on the Smod and Mmod datasets and that our methods perform better. Moreover, our advantage is more pronounced on the Mmod dataset than on the Smod dataset. In particular, our methods significantly outperform the previous best-performing IC-MLNet with a probability of 96.2\%.

\begin{table}
\caption{The Diebold-Mariano test results. The table shows the DM values for the corresponding models versus ours.The normally distributed probabilities corresponding to the DM values are shown in parentheses. }
\label{table: DM}
\footnotesize
\centering
\setlength\tabcolsep{10pt}
\begin{tabular}{|c|c|c|c|}
\hline
Datas & IC-MLNet & MiLeNAS & PC-DARTS \\
\hline
Smod(s)& 1.62(94.7\%) &	0.55(70.7\%) &	0.38(65.0\%)  \\
\hline
Mmod(s) & 1.77(96.2\%) & 0.62(73.2\%) &	0.62(73.2\%) \\
\hline
\end{tabular}

\end{table}

In addition, we find that due to the increasing difficulty in extracting features of Mmod, its performance is less promising in several metrics compared to Smod. This finding is consistent with the previous IC-MLNet. This is evidenced by the fact that Smod significantly outperforms Mmod on the MAE and NSE metrics, and this trend is also observed on other metrics. There are two main reasons for this result: on the one hand, the ensemble forecast has already extracted some features relative to the original data, resulting in information loss; on the other hand, the small number of ensemble forecast members in Mmod makes it impossible to cope with multiple influences. In other words, the multiple random initial conditions in Smod provide more additional information compared to Mmod, which compensates for the information loss to some extent.

\subsection{Precipitation Intensity Classification}

AdaNAS-hss also achieves a surprising performance in all rainfall level criteria in Table \ref{main table}. Our model improves 17.1\% and 8.2\% in ACC and HSS on Smod, 11.6\% and 1.1\% on Mmod, respectively, compared to the previous best-performing IC-MLNet. Besides, our AdaNAS performs better than other NAS methods. The comparison of AdaNAS-mse and AdaNAS-hss shows that introducing HSS into the regularization equation improves both ACC and HSS scores on both datasets.
Next, we will demonstrate a more in-depth study of regularization to explore its impact on ACC and HSS.



\subsection{Regularization Function Analysis}
\label{sec:loss}

\begin{figure}
\centering
\subfigure[Smod]{
\label{fig:subfig:Smod} 
\includegraphics[width=0.20\textwidth]{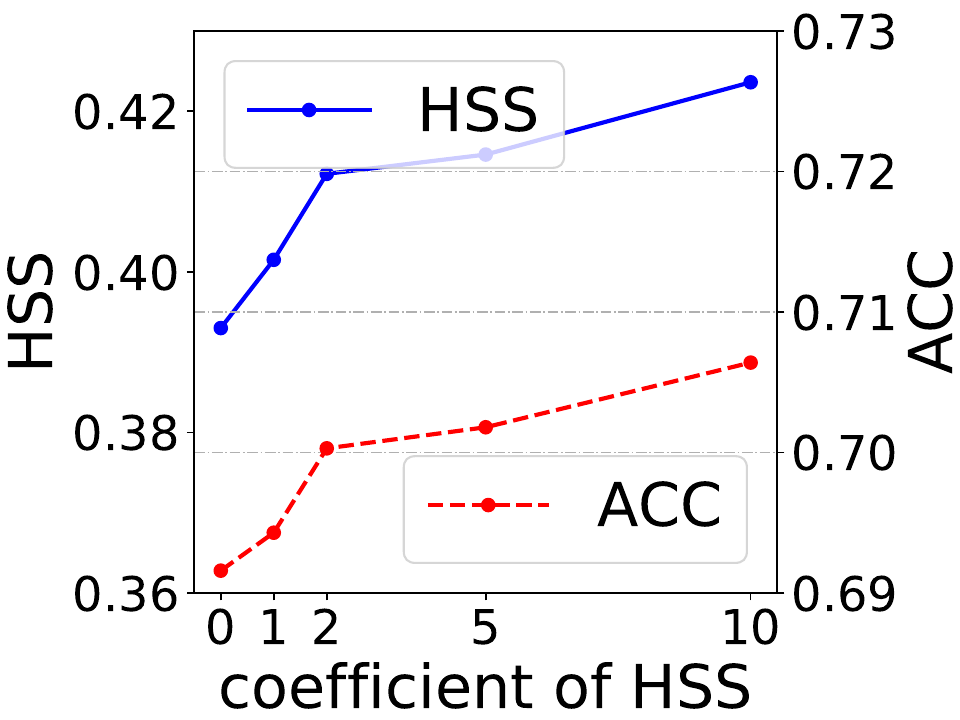}}
\subfigure[Mmod]{
\label{fig:subfig:Mmod} 
\includegraphics[width=0.20\textwidth]{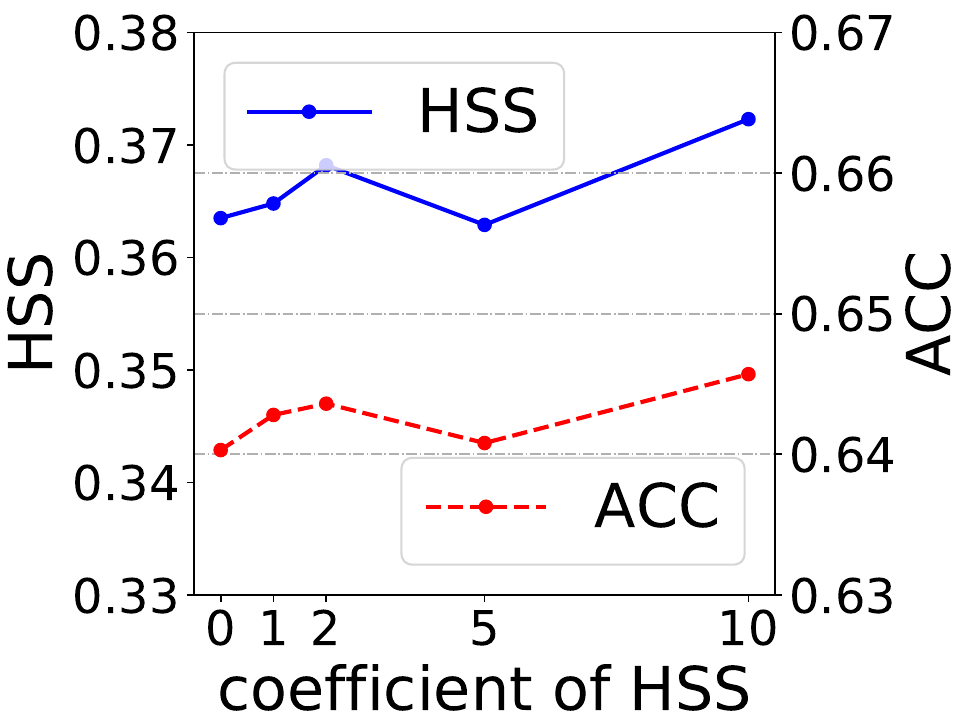}}
\caption{The influence of HSS coefficient $c_H$ in regularization functions.}
\label{loss}
\end{figure}

We further explore the performance of AdaNAS in relation to HSS coefficients in the regularization function. We show model performance for $c_H = 0, 1, 2, 5, $ and 10 in Figure \ref{loss}. In Figure \ref{loss}, 
the model performs best when $c_H = 10$, reaching the highest scores for both ACC and HSS on both Smod and Mmod. This indicates that adding HSS to the regularization function is indeed effective in changing focus and deflecting optimization of model toward precipitation levels, thus improving ACC and HSS. Figure \ref{loss}.(a) shows that HSS and ACC steadily improve and the speed of improvement gradually slows down with the increase of $c_H$ on Smod. However, in Figure \ref{loss}.(b), HSS and ACC boost a little but they are not stable on Mmod, with the increase of $c_H$. It indicates that the HSS coefficients have limited improvement on model, and when $c_H$ exceeds a threshold, the performance of model is almost no longer improved. Besides, in Table \ref{main table}, it can be found that Bias decreases on both Smod and Mmod for $c_H=10$ (AdaNAS-hss) compared to $c_H=0$ (AdaNAS-mse). This indicates that when $c_H$ is greater than 10, it may adversely affect the other precipitation criteria. This may be caused by an excessive deflection of the optimization direction toward precipitation intensity classification.

To study the effect of the HSS coefficients, several sets of experiments with different HSS coefficients are performed, and our experimental results on HSS coefficient show that when the coefficient increases, the bias will be adversely influenced. By observing the penultimate two columns (ACC and HSS) in Table~\ref{tab:hss}, it can be found that the accuracy of precipitation intensity classification gradually improves with the increase of HSS coefficient. The observation of the first column (Bias) shows that the bias achieves the best performance when the HSS coefficient is 0.

\begin{table}
\caption{Results of precipitation amounts and intensity classification with different HSS coefficients on Smod.}
\label{tab:hss}
\centering
\setlength\tabcolsep{3pt}
\begin{tabular}{|c|llll|ll|}

\hline
 & \multicolumn{4}{|c|}{Precipitation amounts}& \multicolumn{2}{|c|}{Levels} \\
\hline
$c_H$ & Bias & MAE &RMSE &NSE &ACC &HSS\\
\hline
10 & 0.8618	&0.9796	&\textbf{2.0420}	&\textbf{0.4460}	&\textbf{0.7064}&	\textbf{0.4236} \\
\hline
5 & 0.8453&	0.9854	&2.0765&	0.4294&	0.7018&	0.4146 \\
\hline
2 & 0.8552&	0.9831	&2.0784	&0.4278&	0.7003&	0.4122 \\
\hline
1 & 0.8656	&\textbf{0.9764}	&2.0490	&0.4438&	0.6943&	0.4015 \\
\hline
0 & \textbf{0.8695}	&0.9917	&2.0754&	0.4305&	0.6916&	0.3930 \\

\hline
\end{tabular}

\end{table}

\subsection{Precipitation Distribution Analysis}
\label{sec:PDA}




\begin{figure}[ht]
\begin{center}
\subfigure[RMSE of precipitation amounts.]{
\label{fig:subfig:RMSE} 
\includegraphics[width=0.9\linewidth]{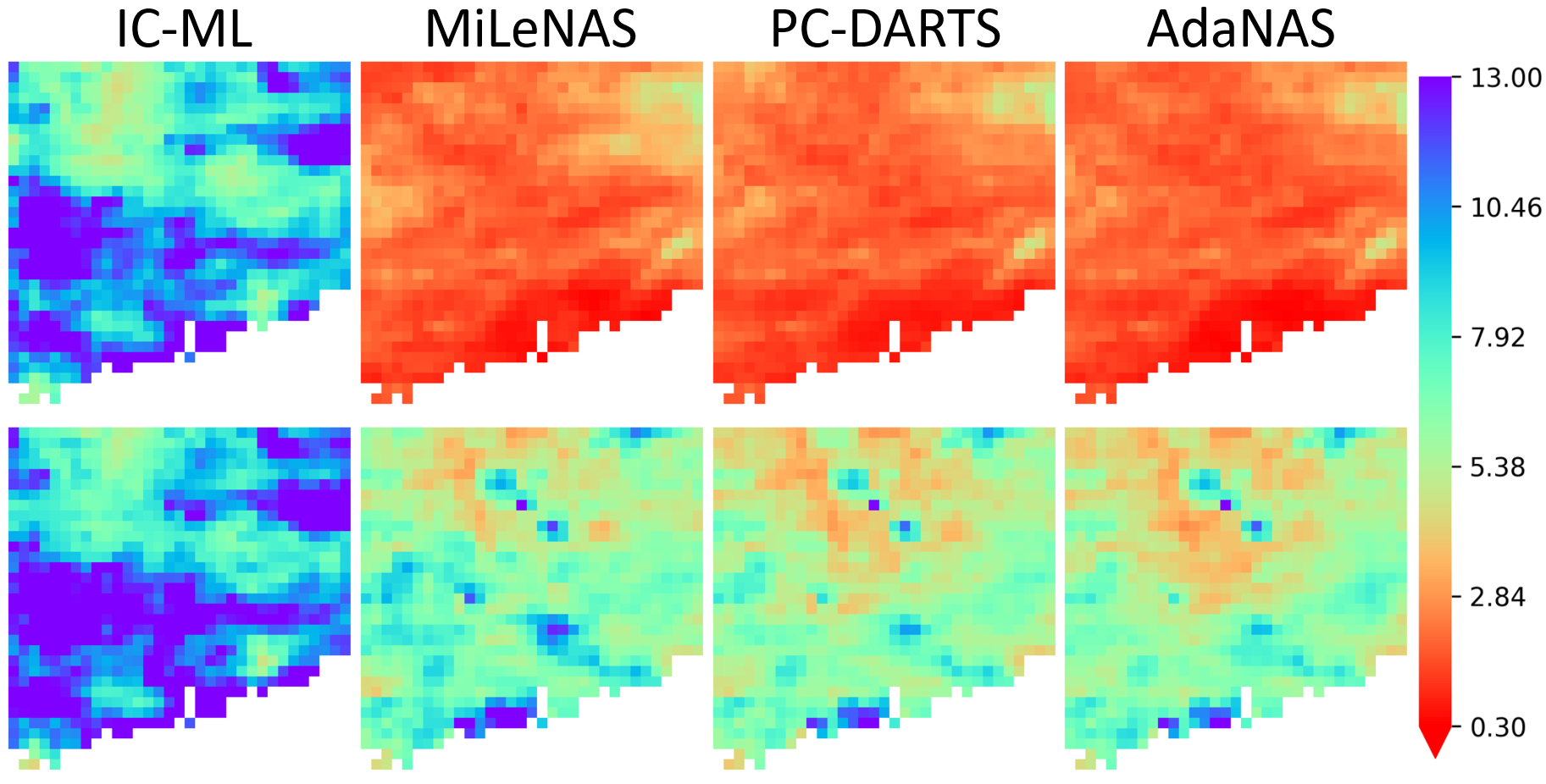}}
\subfigure[MAE of precipitation amounts.]{
\label{fig:subfig:MAE} 
\includegraphics[width=0.9\linewidth]{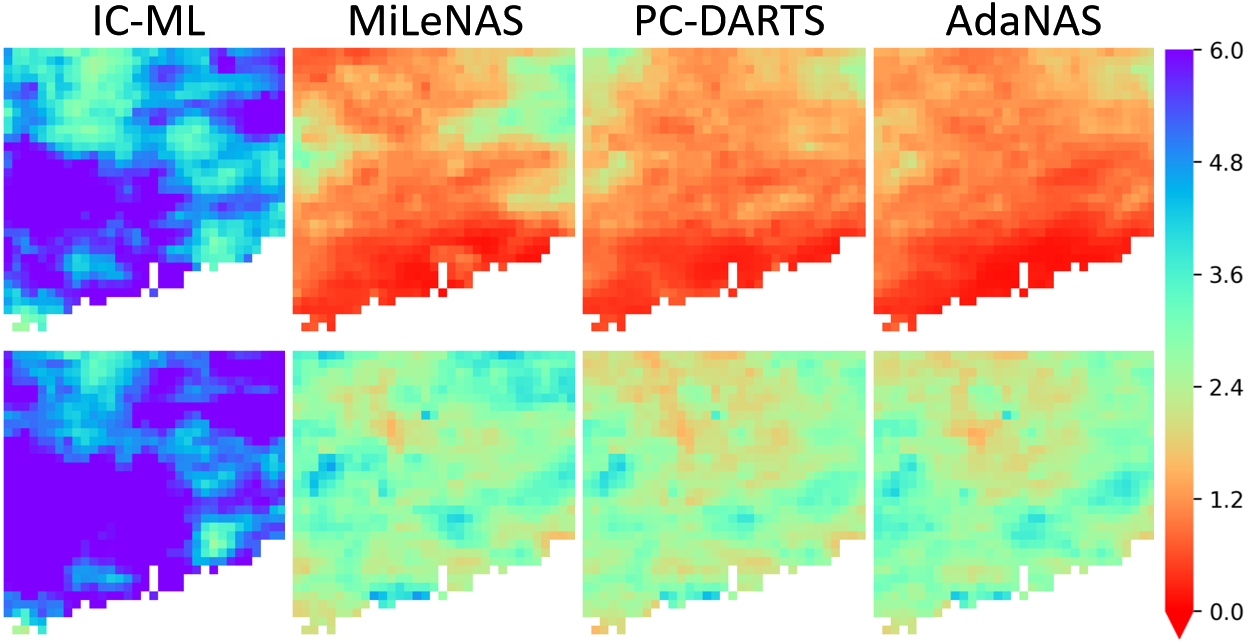}}
\subfigure[ACC of intensity classification.]{
\label{fig:subfig:ACC} 
\includegraphics[width=0.9\linewidth]{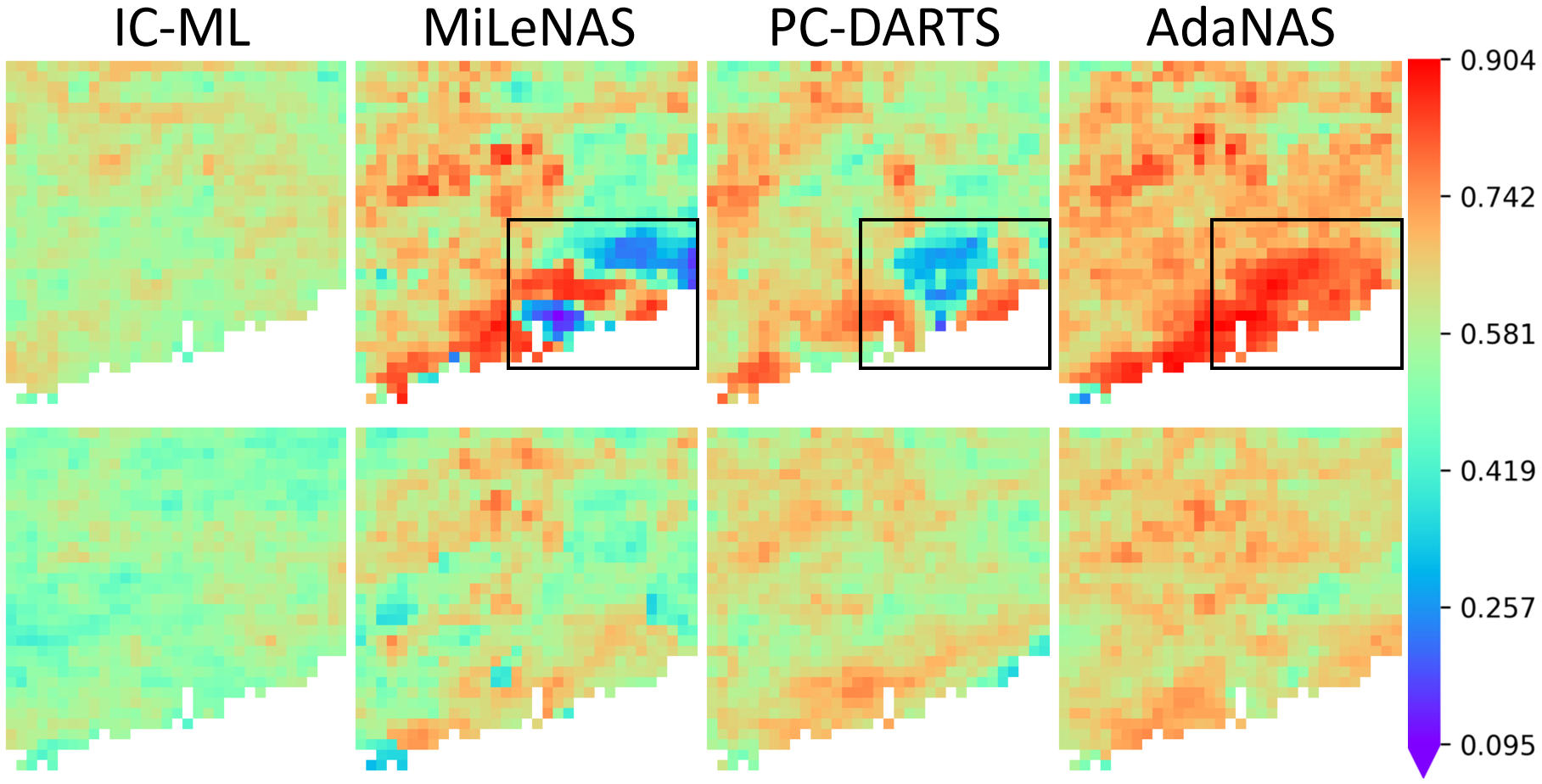}}
\subfigure[HSS of intensity classification.]{
\label{fig:subfig:HSS} 
\includegraphics[width=0.9\linewidth]{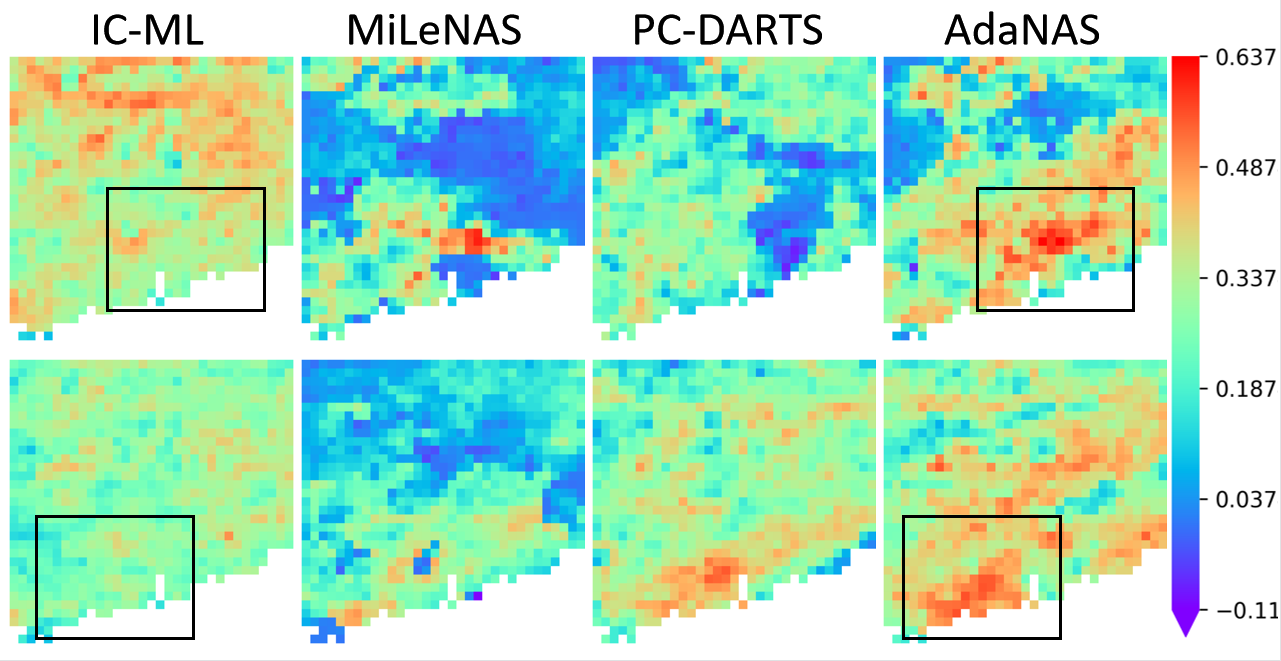}}
\end{center}
   \caption{The distribution of precipitation prediction. The first and second rows are the results on Smod and Mmod, respectively. The vertical coordinates of figures (a),(b), (c) and (d) represent RMSE, MAE, ACC and HSS respectively. The color from blue to red indicates the effect from bad to good.}
\label{fig:mae}
\end{figure}


  We also compared the MAE and HSS performance of AdaNAS with the other two NAS methods and IC-ML by visualizing the spatial distribution of precipitation. The results reveal that our AdaNAS performs significantly better than other methods, both in precipitation prediction and intensity classification. Figure~\ref{fig:subfig:MAE} shows that MAE performance of AdaNAS is significantly better than that of artificially designed IC-ML, especially on Smod. As shown in Figure~\ref{fig:subfig:HSS} and Figure~\ref{fig:subfig:ACC} AdaNAS predicts precipitation intensity (ACC and HSS) more accurately than IC-MLNet and other NAS methods in coastal areas (the areas in the black box). This suggests that CAB and SAB enhance the predictive capability of our method for heavy rainfall areas (The average precipitation is greater in coastal areas compared to inland.)

\subsection{Ablation Study}

\begin{table}
\caption{Ablation experiments of operations.}
\label{operation}
\footnotesize
\centering
\setlength\tabcolsep{2pt}
\begin{tabular}{|c|c|c|c|c|c|c|}
\hline
Operation & Bias & MAE &RMSE &NSE &ACC &HSS \\
\hline
DARTS Ops& 0.8495 &	1.0048&	2.1192&	0.4098	&0.6845	&0.3826 \\
\hline
RB & 0.8651 &	0.9916&	2.0792&	0.4275	&0.6748	&0.3607 \\
\hline
RB\&SAB & 0.8323	&1.0519&	2.1158&	0.4026&	0.6821&	0.3847 \\
\hline
RB\&CAB & 0.8330&	\textbf{0.9860}&	2.0875&	0.4204&	0.6884	&0.3917 \\
\hline
3 Ops & \textbf{0.8695}	& 0.9917	& \textbf{2.0754} &	\textbf{0.4305} &	\textbf{0.6916} &	\textbf{0.3930} \\
\hline
\end{tabular}

\end{table}

To confirm that our proposed rainfall-aware search space can indeed improve the performance of the model, we design five sets of operations for ablation experiments: (1) DARTS Ops~\cite{liu2018darts} (2)RB; (3)RB\&SAB; (4) RB\&CAB; (5)all 3 Ops. As shown in Table \ref{operation}, our method containing all three operations (3 Ops) performs best on all evaluation criteria except MAE. We improved 4.5\%, 5.7\%, 1.9\%, 6.9\%, 1.4\%, 2.2\% on Bias, MAE, RMSE, NSE, ACC, HSS evaluation criteria respectively than using RB\&SAB operation. In particular, the comparison of the first line and the fifth line shows that our search space performs better than that of DARTS when only the operations in the search space are replaced. By comparing the third and fourth rows, we can find that replacing SAB with our CAB leads to an improvement of all evaluation criteria with 0.08\%, 6.3\%, 1.3\%, 4.4\%, 0.9\%, 1.8\%, proving that the proposed CAB can capture more information in ensemble forecasts than SAB. In addition, by comparing the fourth and fifth rows, we can find that using SAB causes all metrics except MAE to improve, which means that SAB can capture some information that CAB fails to capture.

\begin{table}
\caption{Ablation experiments of self-supervised search.}
\label{self-sup}
\footnotesize
\centering
\setlength\tabcolsep{2pt}
\begin{tabular}{|c|c|c|c|c|c|c|}
\hline
Strategy & Bias & MAE &RMSE &NSE &ACC &HSS \\
\hline
Rand. & 0.8482&	1.0961&	2.2196&	0.4098	&0.6776	&0.3783 \\
\hline
Sup. & 0.8665&	1.0186&	2.1696&	0.3776	&0.6807	&0.3790 \\
\hline
Self-sup. & \textbf{0.8695}	&\textbf{0.9917}&	\textbf{2.0754}&	\textbf{0.4305}&	\textbf{0.6916}&	\textbf{0.3930} \\
\hline
\end{tabular}

\end{table}

To demonstrate the effectiveness of self-supervised search, we compare it with random search and supervised search, which uses observed data as labels to supervise the search process. The experimental results are shown in Table \ref{self-sup}, and we can see that self-supervised search outperforms supervised search for all evaluation criteria on Smod, with an especially improvement of 14.0\% on NSE. The reason is that self-supervised search can unbind the initialized network structure and have a broader and freer search space.

\subsection{Time Consumption}
\label{sec: time}

In this section, we test the inference time consumption of the four methods. We conduct experiments on a V100 GPU and test the time required by each of the four methods to validate all samples in the Smod and Mmod validation sets. We set the batch size of the tests to 20. Experimental results show that our method is significantly faster than other methods, especially on the Mmod dataset, as shown in TABLE \ref{table: time}. Specifically, compared to IC-MLNet, the inference speed of our method is improved by 5.7X and 14.4X on the Smod and Mmod datasets, respectively.

\begin{table}
\caption{Time consumption for evaluation.}
\label{table: time}
\footnotesize
\centering
\setlength\tabcolsep{8pt}
\begin{tabular}{|c|c|c|c|c|}
\hline
Methods & IC-MLNet & MiLeNAS & PC-DARTS & AdaNAS\\
\hline
Smod(s)& 33.74&	18.3 &	19.04 &	\textbf{5.95} \\
\hline
Mmod(s) & 21.61& 2.10 &	1.82 &	\textbf{1.50} \\
\hline
\end{tabular}

\end{table}

%% file: Conclusion.tex
\section{Conclusion}
In this work, we propose a novel AdaNAS, which can adaptively design efficient network architectures without excessive manual effort, for the customization of precipitation forecast. We design a rainfall-aware search space and a persistence-aware regularization function respectively for the searching and training process, which improves the accuracy of precipitation forecasting especially in coastal areas. The stable and consistent performance of our AdaNAS on all evaluation criteria outperforms current advanced methods on the large-scale precipitation benchmark TIGGE. 

%% file: Acknowledgements.tex
\section*{Acknowledgements}
This research was supported by the Natural Science Foundation of China under Grant No. U1811464, and was also supported in part by the Guangdong Natural Science Foundation under Grant No. 2018B030312002, in part by the Program for Guangdong Introducing Innovative and Entrepreneurial Teams under Grant NO. 2016ZT06D211, in part by the CCF-Baidu Open Fund OF2021032.

%% file: Appendices.tex
\begin{appendices}
\section{Experimental Setup}
\label{secA1}
The main experimental hyperparameter Settings are shown in the following Table~\ref{hyperparameter}:
\begin{table}[h]
\caption{Main experimental hyperparameter Settings.}
\label{hyperparameter}
\centering
\begin{tabular}{|c|c|}
\hline
hyperparameter & value \\
\hline
search learning rate  & 0.00001 \\
\hline
search batch size & 8 \\
\hline
retrain learning rate & 0.00025 \\
\hline
retrain batch size & 64 \\
\hline
momentum & 0.99 \\
\hline
num of block & 4 \\
\hline
search epochs & 24 \\
\hline
retrain epochs & 300 \\
\hline
\end{tabular}

\vspace{-5mm}
\end{table}

\end{appendices}

%% file: main.bbl
\begin{thebibliography}{10}
\providecommand{\url}[1]{#1}
\csname url@samestyle\endcsname
\providecommand{\newblock}{\relax}
\providecommand{\bibinfo}[2]{#2}
\providecommand{\BIBentrySTDinterwordspacing}{\spaceskip=0pt\relax}
\providecommand{\BIBentryALTinterwordstretchfactor}{4}
\providecommand{\BIBentryALTinterwordspacing}{\spaceskip=\fontdimen2\font plus
\BIBentryALTinterwordstretchfactor\fontdimen3\font minus
  \fontdimen4\font\relax}
\providecommand{\BIBforeignlanguage}[2]{{%
\expandafter\ifx\csname l@#1\endcsname\relax
\typeout{** WARNING: IEEEtran.bst: No hyphenation pattern has been}%
\typeout{** loaded for the language `#1'. Using the pattern for}%
\typeout{** the default language instead.}%
\else
\language=\csname l@#1\endcsname
\fi
#2}}
\providecommand{\BIBdecl}{\relax}
\BIBdecl

\bibitem{zhao2021hourly}
Q.~Zhao, Y.~Liu, W.~Yao, and Y.~Yao, ``Hourly rainfall forecast model using
  supervised learning algorithm,'' \emph{IEEE Transactions on Geoscience and
  Remote Sensing}, vol.~60, pp. 1--9, 2021.

\bibitem{michalakes2001development}
J.~Michalakes, S.~Chen, J.~Dudhia, L.~Hart, J.~Klemp, J.~Middlecoff, and
  W.~Skamarock, ``Development of a next-generation regional weather research
  and forecast model,'' in \emph{Developments in teracomputing}, 2001, pp.
  269--276.

\bibitem{tang2013benefits}
Y.~Tang, H.~W. Lean, and J.~Bornemann, ``The benefits of the met office
  variable resolution nwp model for forecasting convection,''
  \emph{Meteorological Applications}, vol.~20, no.~4, pp. 417--426, 2013.

\bibitem{seity2011arome}
Y.~Seity, P.~Brousseau, S.~Malardel, G.~Hello, P.~B{\'e}nard, F.~Bouttier,
  C.~Lac, and V.~Masson, ``The arome-france convective-scale operational
  model,'' \emph{Monthly Weather Review}, vol. 139, no.~3, pp. 976--991, 2011.

\bibitem{ebert2001ability}
E.~E. Ebert, ``Ability of a poor man's ensemble to predict the probability and
  distribution of precipitation,'' \emph{Monthly Weather Review}, vol. 129,
  no.~10, pp. 2461--2480, 2001.

\bibitem{zhi2011multi}
X.~Zhi, W.~Zhou, and Z.~Xu, ``Multi-model ensemble forecasts of the tc tracks
  over the western pacific using the tigge dataset,'' in \emph{The 3rd
  International Conference on Information Science and Engineering}, 2011, pp.
  4173--4176.

\bibitem{dai2016situation}
K.~Dai, Y.~Cao, Q.~Qian, S.~Gao, S.~Zhao, Y.~Chen, and C.~Qian, ``Situation and
  tendency of operational technologies in short-and medium-range weather
  forecast,'' \emph{Meteor Mon}, vol.~42, no.~12, pp. 1445--1455, 2016.

\bibitem{cuo2011review}
L.~Cuo, T.~C. Pagano, and Q.~Wang, ``A review of quantitative precipitation
  forecasts and their use in short-to medium-range streamflow forecasting,''
  \emph{Journal of hydrometeorology}, vol.~12, no.~5, pp. 713--728, 2011.

\bibitem{gneiting2014probabilistic}
T.~Gneiting and M.~Katzfuss, ``Probabilistic forecasting,'' \emph{Annual Review
  of Statistics and Its Application}, vol.~1, pp. 125--151, 2014.

\bibitem{schaake2007hepex}
J.~C. Schaake, T.~M. Hamill, R.~Buizza, and M.~Clark, ``Hepex: the hydrological
  ensemble prediction experiment,'' \emph{Bulletin of the American
  Meteorological Society}, vol.~88, no.~10, pp. 1541--1548, 2007.

\bibitem{kong2006multiresolution}
F.~Kong, K.~K. Droegemeier, and N.~L. Hickmon, ``Multiresolution ensemble
  forecasts of an observed tornadic thunderstorm system. part i: Comparsion of
  coarse-and fine-grid experiments,'' \emph{Monthly weather review}, vol. 134,
  no.~3, pp. 807--833, 2006.

\bibitem{zhang2021machine}
Y.~Zhang and A.~Ye, ``Machine learning for precipitation forecasts
  postprocessing: Multimodel comparison and experimental investigation,''
  \emph{Journal of Hydrometeorology}, vol.~22, no.~11, pp. 3065--3085, 2021.

\bibitem{li2022convolutional}
W.~Li, B.~Pan, J.~Xia, and Q.~Duan, ``Convolutional neural network-based
  statistical post-processing of ensemble precipitation forecasts,''
  \emph{Journal of hydrology}, vol. 605, p. 127301, 2022.

\bibitem{ghazvinian2021novel}
M.~Ghazvinian, Y.~Zhang, D.-J. Seo, M.~He, and N.~Fernando, ``A novel hybrid
  artificial neural network-parametric scheme for postprocessing medium-range
  precipitation forecasts,'' \emph{Advances in Water Resources}, vol. 151, p.
  103907, 2021.

\bibitem{xu2021multi}
F.~Xu, G.~Li, Y.~Du, Z.~Chen, and Y.~Lu, ``Multi-layer networks for ensemble
  precipitation forecasts postprocessing,'' in \emph{Proceedings of the AAAI
  Conference on Artificial Intelligence}, vol.~35, no.~17, 2021, pp.
  14\,966--14\,973.

\bibitem{duan2019handbook}
Q.~Duan, F.~Pappenberger, A.~Wood, H.~L. Cloke, and J.~C. Schaake,
  \emph{Handbook of hydrometeorological ensemble forecasting}.\hskip 1em plus
  0.5em minus 0.4em\relax Springer Berlin/Heidelberg, Germany, 2019, vol.~10.

\bibitem{vannitsem2021statistical}
S.~Vannitsem, J.~B. Bremnes, J.~Demaeyer, G.~R. Evans, J.~Flowerdew, S.~Hemri,
  S.~Lerch, N.~Roberts, S.~Theis, A.~Atencia \emph{et~al.}, ``Statistical
  postprocessing for weather forecasts: Review, challenges, and avenues in a
  big data world,'' \emph{Bulletin of the American Meteorological Society},
  vol. 102, no.~3, pp. E681--E699, 2021.

\bibitem{vannitsem2018statistical}
S.~Vannitsem, D.~S. Wilks, and J.~Messner, \emph{Statistical postprocessing of
  ensemble forecasts}.\hskip 1em plus 0.5em minus 0.4em\relax Elsevier, 2018.

\bibitem{he2021automl}
X.~He, K.~Zhao, and X.~Chu, ``Automl: A survey of the state-of-the-art,''
  \emph{Knowledge-Based Systems}, vol. 212, p. 106622, 2021.

\bibitem{xu2019pc}
Y.~Xu, L.~Xie, X.~Zhang, X.~Chen, G.-J. Qi, Q.~Tian, and H.~Xiong, ``Pc-darts:
  Partial channel connections for memory-efficient architecture search,''
  \emph{arXiv preprint arXiv:1907.05737}, 2019.

\bibitem{he2020milenas}
C.~He, H.~Ye, L.~Shen, and T.~Zhang, ``Milenas: Efficient neural architecture
  search via mixed-level reformulation,'' in \emph{Proceedings of the IEEE/CVF
  Conference on Computer Vision and Pattern Recognition}, 2020, pp.
  11\,993--12\,002.

\bibitem{ahmed2018maskconnect}
K.~Ahmed and L.~Torresani, ``Maskconnect: Connectivity learning by gradient
  descent,'' in \emph{Proceedings of the European Conference on Computer Vision
  (ECCV)}, 2018, pp. 349--365.

\bibitem{you2020greedynas}
S.~You, T.~Huang, M.~Yang, F.~Wang, C.~Qian, and C.~Zhang, ``Greedynas: Towards
  fast one-shot nas with greedy supernet,'' in \emph{Proceedings of the
  IEEE/CVF Conference on Computer Vision and Pattern Recognition}, 2020, pp.
  1999--2008.

\bibitem{cai2018proxylessnas}
H.~Cai, L.~Zhu, and S.~Han, ``Proxylessnas: Direct neural architecture search
  on target task and hardware,'' \emph{arXiv preprint arXiv:1812.00332}, 2018.

\bibitem{tan2019efficientnet}
M.~Tan and Q.~Le, ``Efficientnet: Rethinking model scaling for convolutional
  neural networks,'' in \emph{International Conference on Machine
  Learning}.\hskip 1em plus 0.5em minus 0.4em\relax PMLR, 2019, pp. 6105--6114.

\bibitem{liu2018darts}
H.~Liu, K.~Simonyan, and Y.~Yang, ``Darts: Differentiable architecture
  search,'' \emph{arXiv preprint arXiv:1806.09055}, 2018.

\bibitem{yu2019evaluating}
K.~Yu, C.~Sciuto, M.~Jaggi, C.~Musat, and M.~Salzmann, ``Evaluating the search
  phase of neural architecture search,'' \emph{arXiv preprint
  arXiv:1902.08142}, 2019.

\bibitem{yang2019evaluation}
A.~Yang, P.~M. Esperan{\c{c}}a, and F.~M. Carlucci, ``Nas evaluation is
  frustratingly hard,'' \emph{arXiv preprint arXiv:1912.12522}, 2019.

\bibitem{zela2020bench}
A.~Zela, J.~Siems, and F.~Hutter, ``Nas-bench-1shot1: Benchmarking and
  dissecting one-shot neural architecture search,'' \emph{arXiv preprint
  arXiv:2001.10422}, 2020.

\bibitem{li2020improving}
X.~Li, C.~Lin, C.~Li, M.~Sun, W.~Wu, J.~Yan, and W.~Ouyang, ``Improving
  one-shot nas by suppressing the posterior fading,'' in \emph{Proceedings of
  the IEEE/CVF Conference on Computer Vision and Pattern Recognition}, 2020,
  pp. 13\,836--13\,845.

\bibitem{moons2020distilling}
B.~Moons, P.~Noorzad, A.~Skliar, G.~Mariani, D.~Mehta, C.~Lott, and
  T.~Blankevoort, ``Distilling optimal neural networks: Rapid search in diverse
  spaces,'' \emph{arXiv preprint arXiv:2012.08859}, 2020.

\bibitem{li2020block}
C.~Li, J.~Peng, L.~Yuan, G.~Wang, X.~Liang, L.~Lin, and X.~Chang,
  ``Block-wisely supervised neural architecture search with knowledge
  distillation,'' in \emph{Proceedings of the IEEE/CVF Conference on Computer
  Vision and Pattern Recognition}, 2020, pp. 1989--1998.

\bibitem{He2020MomentumCF}
K.~He, H.~Fan, Y.~Wu, S.~Xie, and R.~B. Girshick, ``Momentum contrast for
  unsupervised visual representation learning,'' \emph{2020 IEEE/CVF Conference
  on Computer Vision and Pattern Recognition (CVPR)}, pp. 9726--9735, 2020.

\bibitem{rumelhart1986learning}
D.~E. Rumelhart, G.~E. Hinton, and R.~J. Williams, ``Learning representations
  by back-propagating errors,'' \emph{nature}, vol. 323, no. 6088, pp.
  533--536, 1986.

\bibitem{Resnet}
K.~He, X.~Zhang, S.~Ren, and J.~Sun, ``Deep residual learning for image
  recognition,'' in \emph{2016 IEEE Conference on Computer Vision and Pattern
  Recognition (CVPR)}.\hskip 1em plus 0.5em minus 0.4em\relax Las Vegas, NV,
  United States: IEEE, 2016, pp. 770--778.

\bibitem{yang2021simam}
L.~Yang, R.-Y. Zhang, L.~Li, and X.~Xie, ``Simam: A simple, parameter-free
  attention module for convolutional neural networks,'' in \emph{International
  Conference on Machine Learning}.\hskip 1em plus 0.5em minus 0.4em\relax PMLR,
  2021, pp. 11\,863--11\,874.

\bibitem{vaswani17nips}
A.~Vaswani, N.~Shazeer, N.~Parmar, J.~Uszkoreit, L.~Jones, A.~N. Gomez,
  L.~Kaiser, and I.~Polosukhin, ``Attention is all you need,'' in \emph{nips},
  2017.

\bibitem{NSAS}
M.~Zhang, H.~Li, S.~Pan, X.~Chang, and S.~Su, ``Overcoming multi-model
  forgetting in one-shot nas with diversity maximization,'' in
  \emph{Proceedings of the IEEE/CVF Conference on Computer Vision and Pattern
  Recognition}, 2020, pp. 7809--7818.

\bibitem{diebold1995paring}
F.~X. Diebold and R.~S. Mariano, ``Com paring predictive accu racy,''
  \emph{Journal of Business and Economic Statistics}, vol.~13, no.~3, pp.
  253--263, 1995.

\end{thebibliography}
